\documentclass[letterpaper, 10 pt, conference]{ieeeconf}  % Comment this line out if you need a4paper

\IEEEoverridecommandlockouts

\usepackage[linesnumbered,ruled,vlined]{algorithm2e}

\usepackage{color}
\definecolor{darkred}{RGB}{205,38,38}

\usepackage[colorlinks=true,
linkcolor=darkred,       
anchorcolor=blue, 
citecolor=darkred,        
filecolor=darkred, 
urlcolor=blue,
]{hyperref}

\usepackage{caption}
\usepackage{subfig}
\usepackage{float} 
\let\labelindent\relax 
\usepackage{enumitem}
\usepackage{booktabs} 
\usepackage{multirow}
\usepackage{url}
\usepackage{cite}
\usepackage{array}
\newcolumntype{L}[1]{>{\raggedright\let\newline\\\arraybackslash\hspace{0pt}}m{#1}}
\newcolumntype{C}[1]{>{\centering\let\newline\\\arraybackslash\hspace{0pt}}m{#1}}
\newcolumntype{R}[1]{>{\raggedleft\let\newline\\\arraybackslash\hspace{0pt}}m{#1}}
\usepackage{threeparttable}
\usepackage{makecell}
\usepackage{balance}
\usepackage{amsmath}
\usepackage{amssymb,bm}
\usepackage{amsfonts}
\usepackage{enumerate}
\usepackage{graphicx}

\title{\LARGE \bf \textit{P}$^2$ \textit{Explore}: Efficient Exploration in Unknown Cluttered Environment with Floor Plan Prediction}
	\author{Kun Song$^{1,2}$, Gaoming Chen$^1$, Masayoshi Tomizuka$^3$, Wei Zhan$^3$, Zhenhua Xiong$^1$, and Mingyu Ding$^3$
		\thanks{Project page: \href{https://song-kun.github.io/projects/p2explore/}{https://song-kun.github.io/projects/p2explore/}.
		}
		\thanks{$^1$K. Song, G. Chen, and Z. Xiong are with the School of Mechanical Engineering, Shanghai Jiao Tong University, Shanghai, China (e-mail: \{coldtea, cgm1015, mexiong\}@sjtu.edu.cn).
		}
		\thanks{$^2$K. Song is with the Department of Computer Science, The University of Hong Kong, Hong Kong SAR, China.
		}
		\thanks{$^3$M. Tomizuka, W. Zhan, and M. Ding are with the Department of Mechanical Engineering, University of California, Berkely, CA 94720, USA (e-mail: \{tomizuka, wzhan, myding\}@berkeley.edu)
		}
		\thanks{
			Corresponding author: Mingyu Ding.
		}
	}
	
\begin{document}

\maketitle
\thispagestyle{empty}
\pagestyle{empty}

\begin{abstract}
Robot exploration aims at the reconstruction of unknown environments, and it is important to achieve it with shorter paths.
Traditional methods focus on optimizing the visiting order of frontiers based on current observations, which may lead to local-minimal results.
Recently, by predicting the structure of the unseen environment, the exploration efficiency can be further improved. 
However, in a cluttered environment, due to the randomness of obstacles, the ability to predict is weak. 
Moreover, this inaccuracy will lead to limited improvement in exploration.
Therefore, we propose FPUNet which can be efficient in predicting the layout of noisy indoor environments.
Then, we extract the segmentation of rooms and construct their topological connectivity based on the predicted map.
The visiting order of these predicted rooms is optimized which can provide high-level guidance for exploration. 
The FPUNet is compared with other network architectures which demonstrates it is the SOTA method for this task.
Extensive experiments in simulations show that our method can shorten the path length by 2.18\% to 34.60\% compared to the baselines.
\end{abstract}

\section{Introduction}
The primary goal of robotic systems is to perceive the physical world based on observational information and then interact with it, which is common in tasks such as navigation \cite{hoeller2024anymal,yokoyama2024vlfm}, manipulation \cite{padalkar2023open}, and exploration \cite{cao2023representation, zhou2023racer}. 
The physical environments can be classified into two categories based on their initial status: known and unknown.
In fully known environments, it is possible to design algorithms to achieve optimal performance.
For example, utilizing the map of the scene, a robot can find an optimal path from one point to another.
However, in unknown environments, these algorithms are limited to current observations, which easily lead to locally optimal solutions.

\begin{figure}[!t]
	\centering   
	\includegraphics[width=2.7in]{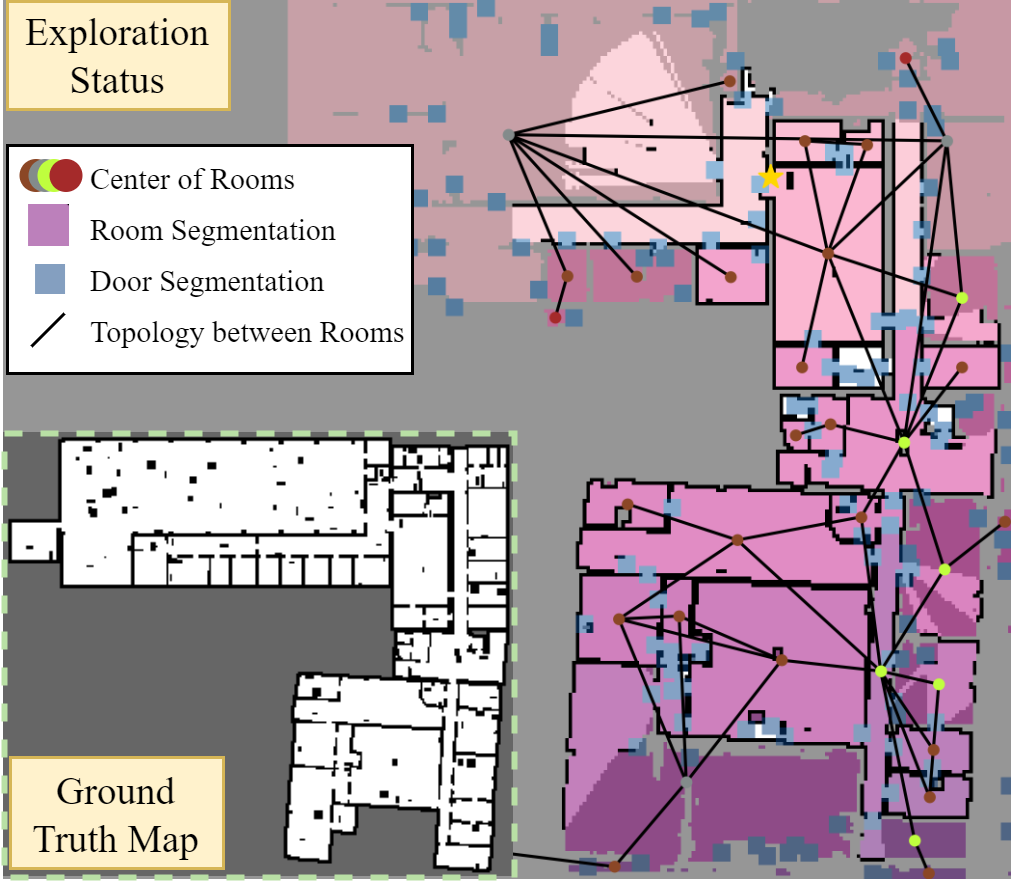}
	\caption{
		The bottom left shows the ground truth of the environment, while the right side displays the exploration status at a specific moment. 
	We first obtain the predicted floor plan using FPUNet.
		Then, the segmentation of rooms (purple area) and their topological connectivity (black line) are extracted to accelerate exploration.}
	\label{fig:fig1}
	\vspace*{-5mm}
\end{figure}

\begin{figure*}[!t]
	\centering   
	\includegraphics[width=6.5in]{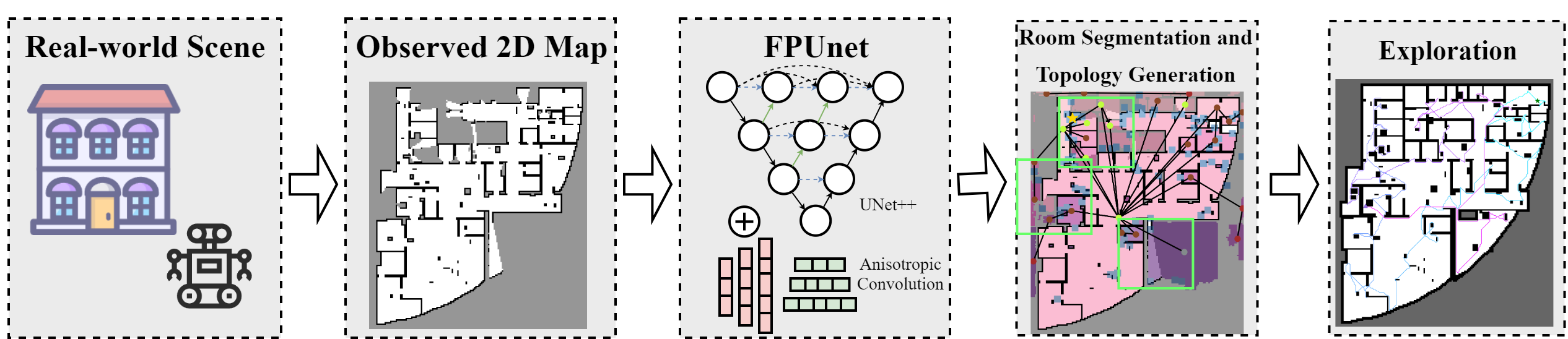}
	\caption{Illustration of the framework of the proposed method. 
		For real-world scenes, there will be various obstacles in the 2D grid map.
		To achieve efficient map prediction, we denoise the 2D map first and then perform prediction based on the floor plan. 
		Finally, the predicted map can be used to extract room segmentation and their topology, which provides guidance for the downstream tasks.}
	\label{fig:total}
	\vspace*{-5mm}
\end{figure*}
Robot exploration, also called active Simultaneous Localization and Mapping (SLAM), aims at actively reconstructing the unknown environment and has been widely studied in recent years, focusing on robot perception and planning\cite{cao2023representation, zhou2023racer, best2024multi, tao2023seer, tao2024learning,katyal2019uncertainty,elhafsi2020map,wei2021occupancy,yokoyama2024vlfm,zheng2025aage}.
To achieve it, most existing methods focus on utilizing the currently observed environment to find the visiting order of frontiers, thus determining the next goal \cite{zhou2023racer, hardouin2023multirobot, umari2017autonomous}.
In \cite{umari2017autonomous}, a greedy algorithm is employed. 
To achieve exploration, the next best view (NBV) is selected as the frontier with maximum utility considering the information gain and navigation cost in the current map.
In \cite{zhou2023racer,hardouin2023multirobot}, the author considers constructing a visiting sequence of all frontiers as a Traveling Salesman Problem (TSP). 
By optimizing the visiting order to find the shortest path, the exploration can be accelerated.
The distributed form of this problem is studied in \cite{patwardhan2024distributed, asgharivaskasi2024riemannian}.
However, these methods focus on the current observed environment, and when new information emerges, these decisions may be suboptimal or even poor, leading to inefficient back-and-forth movement.

Therefore, a question arises: \textbf{in the exploration task, is it possible to find the global shortest trajectory?}
Inspired by the map coverage problem \cite{choset2000coverage}, for a fully known environment, we can find an optimal trajectory to achieve coverage of the environment.
The primary difference between exploration and map coverage lies in whether prior information about the environment is available. 
Therefore, if we can predict the status of the unobserved parts based on a partially observed environment, it would be possible to generate an optimal trajectory for exploration.
Previous works have explored how to achieve map prediction in indoor environments \cite{shrestha2019learned,ericson2021understanding,tao2024learning} which can be categorized into two types: cluttered environments and floor plans.
Floor plans are the structure of walls, while cluttered environments are the combination of walls and various objects.

Some researchers focus on directly predicting the unseen cluttered environments, where the presence of furniture and other objects results in irregular structures on the map. 
However, due to this irregularity, the map prediction is prone to fail in complex scenes.
In \cite{katyal2019uncertainty, elhafsi2020map,katyal2021high}, the robot is equipped with 2D LiDAR to perform map prediction. 
Due to the randomness of obstacles, the effectiveness of the map predictor is poor, offering only a modest acceleration for exploration.
Even when equipped with advanced sensors, such as RGB-D cameras, only the nearby portions of the map can be predicted \cite{ramakrishnan2020occupancy,wang2021learning,sharma2023proxmap,ji2023ddp,chen2023stexplorer}.
Other researchers focus on map prediction in environments with floor plans only.
In \cite{chang2007p, luperto2019predicting}, geometry-based methods are used to achieve the prediction. 
However, traditional methods face challenges in generalization, leading to a greater focus on using deep learning techniques.
In \cite{ericson2022floorgent,ericson2024beyond}, a learning-based method is utilized, and floor plans are predicted using an auto-regressive sequence prediction method which provides satisfying results.
However, most of the scenes in the real world are cluttered.
Therefore, the application of focusing on pure floor plan prediction is limited.

Once the predicted map is obtained, how to utilize it to accelerate exploration is a key question.
As mentioned above, a perfect map predictor does not exist. 
The farther a location is from the observed area, the higher the probability of making an error in predicting its occupancy status.
Therefore, a key research question is how to utilize an imperfect map predictor to accelerate exploration.

In \cite{shrestha2019learned}, NBV is selected based on the predicted floor plans.
The information gain for a certain frontier is evaluated on the predicted map.
In \cite{tao2024learning}, a lightweight neural network (NN) is used as a map predictor in a floor plan environment, and a deep reinforcement learning-based method is proposed for efficient exploration.
However, current methods mainly focus on map prediction in size-limited and noise-free scenes, which rarely appear in the real world, leading to a constrained application.

In conclusion, there are two major challenges in using map prediction to accelerate exploration.
Firstly, how to achieve efficient map prediction in general cluttered environments.
Secondly, how to use inaccurate prediction to accelerate exploration.
In our work, we minimize the gap of current research in predicting maps for cluttered scenarios of any size and how to implement them for exploration robustly. 
We propose a framework called \textit{P}$^2$ \textit{Explore} (Floor \textbf{P}lan \textbf{P}rediction Explore).
For the first challenge, rather than directly predicting the scene with obstacles, we focus on predicting the floor plans that are more feasible to achieve.
For the second one, we extract the segmentation of rooms in the predicted map and use their connectivity to provide high-level guidance for exploration.
To be more specific, firstly, we propose a new architecture of UNet for \textbf{f}loor plan \textbf{p}rediction called FPUNet.
It is trained on a dataset containing real-world scenes.
Then, we combine predictions from different frontiers, enabling the generation of a global predicted map. 
As a result, our algorithm can be extended to maps of any scale. 
Furthermore, we perform room segmentation based on the predicted map, extracting the specific locations of individual rooms and automatically generating the topological connectivity between them.
Then, we optimize the visiting order of all rooms including the predicted ones to accelerate exploration.
The illustration for the proposed method can be found in Fig. \ref{fig:total}.
Our main contributions are:
\begin{itemize}[leftmargin=*]
	\item FPUNet is proposed for map prediction in cluttered environment, which is the state-of-the-art (SOTA) architecture.
	\item Room segmentation and connectivity are generated. Then, the visiting order of all rooms are optimized based one the connectivity to accelerate exploration.
	\item The effectiveness of \textit{P}$^2$ \textit{Explore} is validated in both simulations and real-world environment with noise.
\end{itemize}

\section{Methodology}
In this section, we will first introduce the map predictor and then explain how to implement it for exploration.

\subsection{Map predictor}
We assume that the robot, equipped with 2D sensors such as a 2D LiDAR, is exploring the environment.
At time step $t$, the observed grid map can be denoted as $G(t) \in \mathcal{C}^{H(t)\times W(t)}$, where $\mathcal{C}$ is the set containing status for each location, $H(t)\in \mathbb{Z}^+$ and $W(t) \in \mathbb{Z}^+$ is the time-varying height and width of the map.
There are three different statuses for each cell: occupied, free, and unknown.
A frontier $f_i$ is defined as the boundary between free and unknown spaces.
Local map around $f_i$ is defined as
\begin{equation}
	M(f_i) = \{x\ \big|\ ||x - f_i||_1\leq R, x\in G\},
	\label{equ:localmap}
\end{equation}
where $R$ is the range of local map.

For indoor environments, furniture such as sofas, chairs, and various objects of different shapes located in arbitrary positions create irregular obstacle areas on $M$. 
As a result, it is difficult to perform map prediction directly to get the unseen parts with obstacles. 
Therefore, we consider performing prediction from $M$ to the floor plan $\hat{\mathcal{P}}$.
A lightweight NN $\psi$ is used to perform prediction.
Assuming that the \textbf{fully} observed floor plan around $f_i$ is denoted as $\hat{\mathcal{P}}_{gt}(f_i)$.
For the map predictor $\psi$, our goal is to find parameters $\theta$ of $\psi$ that can minimize the error between the predicted floor plan $\hat{\mathcal{P}}(f_i) = \psi_\theta (M(f_i))$ and the ground-truth $\hat{\mathcal{P}}_{gt}(f_i)$
\begin{equation}
	\arg\min_\theta \mathcal{D}(\hat{\mathcal{P}}(f_i), \hat{\mathcal{P}}_{gt}(f_i)),
\end{equation}
where $\mathcal{D}$ represents a distance metric for two maps.
The detailed information will be presented in Section \ref{sec:train}.

\subsubsection{Dataset Generation}
Similarly with \cite{shrestha2019learned,ericson2024beyond}, we can obtain different scenes from KTH floor plan dataset\cite{aydemir2012can}.
This dataset describes the floor plan in the form of line segments, specifying the start and end points.
Each line segment can be categorized into three types: wall, window, and door. 
We will describe how to generate our training data based on this dataset.

The locations of walls within the environment indicate the traversability of the scene. 
Then, we map the positions of walls in each floor plan to a grid map, where each cell represents 0.2\,m in the real world.
Next, by using the flood-fill algorithm from a point in indoor free spaces, we can identify the interior of each scene.
Therefore, each cell is then assigned a label as either occupied, free, or unknown.
To simulate irregular obstacles in the real world, we add random obstacles of varying sizes and quantities to each map to represent real-world environments.

Then, a robot equipped with a LiDAR with a range of 12\,m is randomly placed at a point in the map. 
We assume the robot performs a virtual exploration based on the NBV strategy \cite{umari2017autonomous}. 
Each time, the robot selects a frontier $f_i$ as the goal for exploration. 
At this point, the local map $M(f_i)$ around $f_i$ can be obtained.
Additionally, we can also acquire the ground truth fully observed floor plan $\hat{\mathcal{P}}_{gt}(f_i)$. 
Thus, we can obtain pairs like $(M(f_i), \hat{\mathcal{P}}_{gt}(f_i))$.
In different scenes and different steps of exploration, we can obtain similar data, which constitutes our dataset.

Due to that in the KTH dataset, similar floor plans may appear in different maps.
For example, they could be different floors in the same building.
Therefore, if we shuffle the entire dataset, training data and testing data will share some similar floor plans.
To avoid this, we remove those similar plans and obtain 140 different floor plans.
The average and maximum size of the scene is 745.59\,m$^2$ and 4548.48\,m$^2$.

\subsubsection{Training Details}
\label{sec:train}
For the original dataset, we divide it into training, testing, and validation sets according to different scenes, with proportions of 70\%, 15\%, and 15\%, respectively. 
In Eq. \ref{equ:localmap}, $R$ is set to 12.8\,m, which makes the input for the network is 128$\times$128.

We transform the original grid map into a three-channel image, where the three channels represent the status of free, occupied, and unknown spaces, respectively. 
Therefore, the input dimension of our model is 3$\times$128$\times$128.
The network outputs the probability of each cell in the map belonging to each of the three categories in the predicted result.

In order to increase the map prediction accuracy for this task, we propose a new architecture of NN for \textbf{f}loor plan \textbf{p}rediction based on \textbf{UNet} called \textbf{FPUNet}.
Since this task needs to consider both low-frequency floor plan information and high-frequency noise, we choose UNet++ \cite{zhou2018unet++} as the backbone, which captures features at different depth levels. 
In addition, the floor plan contains various structures like lines, rectangles, and semicircles, which result in anisotropic map features. 
Therefore, we introduce anisotropic convolution \cite{Li2022anisConvo} into our network. 
First, we use unidirectional convolutional kernels with learnable weight factors, which enhance the network's ability to capture floor plan features in different directions.
Second, we incorporate convolutional kernels of different scales into our network to improve its ability to capture information at various spatial ranges.

Since the number of data for three categories in the map is different, and in exploration tasks, the location of walls represented by the status of occupied plays a crucial role in separating different rooms.
Therefore, accurately predicting the positions of walls is essential. 
We use weighted cross-entropy as our loss function and apply a higher weight to the occupied category.
The loss for a certain pair is
\begin{equation}
    L=\sum_n l_n,\ l_n = -w_{x_n} \log (\hat{x}_n),
    \label{equ:loss}
\end{equation}
where $l_n$ is the loss for a certain pixel $n$, $x_n$ is the label for $n$, $\hat{x}_n$ is the predicted probability of $n$ belonging to $x_n$.
$w_{x_n}$ is the weight for class $x_n$.

We set the optimizer to Adam with an initial learning rate of 0.0002, adjust using the cosine annealing strategy, train for 400 epochs, and save the network that performs best on the validation set as the final result.

\subsubsection{Obtain the Global Predicted Map}
In this section, we will introduce how to fuse local predicted maps of fixed size into a global map. 
First, for each pixel in the predicted result, we define its category as the one with the highest probability. 
Then, we turn the value in this map to 0-255, ensuring that it can later be fused with the already observed map.

To enable map prediction for environments of any size, we use the local map around each frontier as input to obtain the local predicted map.
Assuming that in the current partially observed map, we can obtain some cluttered frontiers $f_i,i=1\cdots n$.
Local maps around $f_i$ is $M(f_i)$.
Based on the trained map predictor $\psi_\theta$, we can obtain the predicted map $\hat{P}(f_i)$.
The detailed algorithm for merging the local predicted maps can be found in Algorithm \ref{alg:merge_map}.
For each local predicted map $\hat{P}(f_i)$, the valid prediction information (occupied or traversable) is added to the corresponding location in the global predicted map $\hat{G}$. 
Additionally, we retain the already observed information in the original map.
\begin{algorithm}
	\KwIn{original map $X$, frontiers $\{f_i\}$, predicted local maps $\{\hat{P}(f_i)\}$} 
	\KwOut{predicted global map $\hat{G}$}
	$\hat{G}$ $\leftarrow$ $G$\\
	$\text{Unknown\_Index}$ $\leftarrow$ $\texttt{where}(G==\text{uknown})$\\
	\For{\rm $\hat{P}(f_i) \in \{\hat{P}(f_i)\}$}
	{
		$\text{Predicted\_Index}$ $\leftarrow$ $\texttt{where}(\hat{P}(f_i)\neq \text{uknown})$\\
		$\text{Index}$ $\leftarrow$ $\text{Unknown\_Index} \cap \text{Predicted\_Index}$\\
		$\hat{G}[\text{Index}]$ $\leftarrow$ $\hat{P}(f_i) [\text{Index}]$
	}
	\caption{Merge Local Prediction}
	\label{alg:merge_map} 
\end{algorithm}

\subsection{Implementation of Map Predictor in Exploration}
Since the map predictor is imperfect, it is essential to find a robust algorithm that can effectively utilize the predicted results to accelerate exploration, especially in indoor cluttered environments.
For the task of map prediction, it is challenging to predict the exact size and structure of a room. 
However, predicting the existence of an unobserved room and its connectivity with other rooms is more feasible. 
Therefore, we propose \textit{P$^2$ Explore} algorithm based on the predicted room existence and connectivity.

\subsubsection{Room Segmentation and Topology Generation}
In this section, we abstract the predicted information by extracting segmentation of rooms and the topological connectivity between them from the predicted map $\hat{G}$. 

\begin{figure}[!t]
	\centering  
	\subfloat[]{ 
		\centering    
		\includegraphics[width=1.3in]{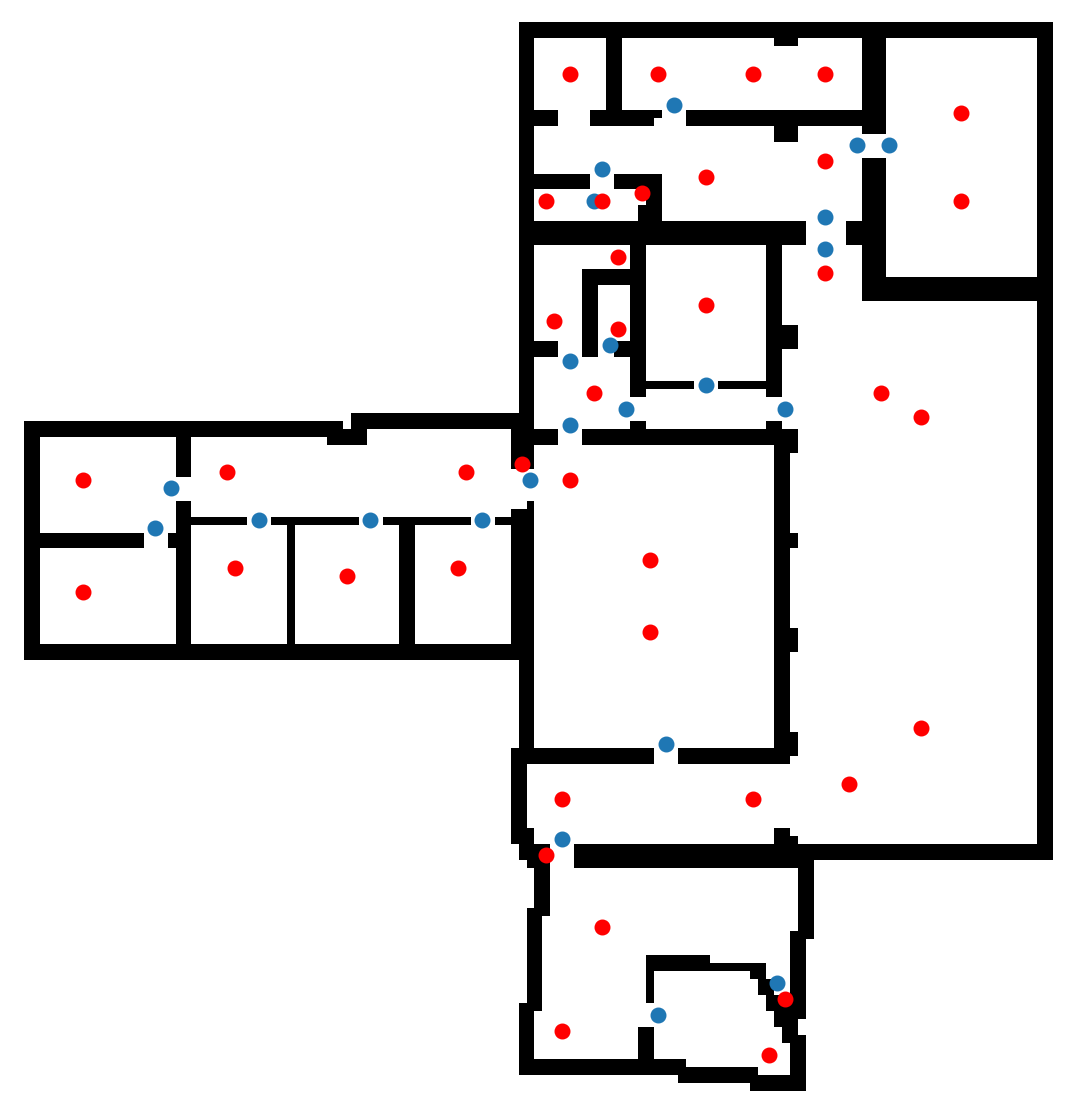}
		\label{fig:room_door}
	}
	\subfloat[]{   
		\centering    
		\includegraphics[width=1.3in]{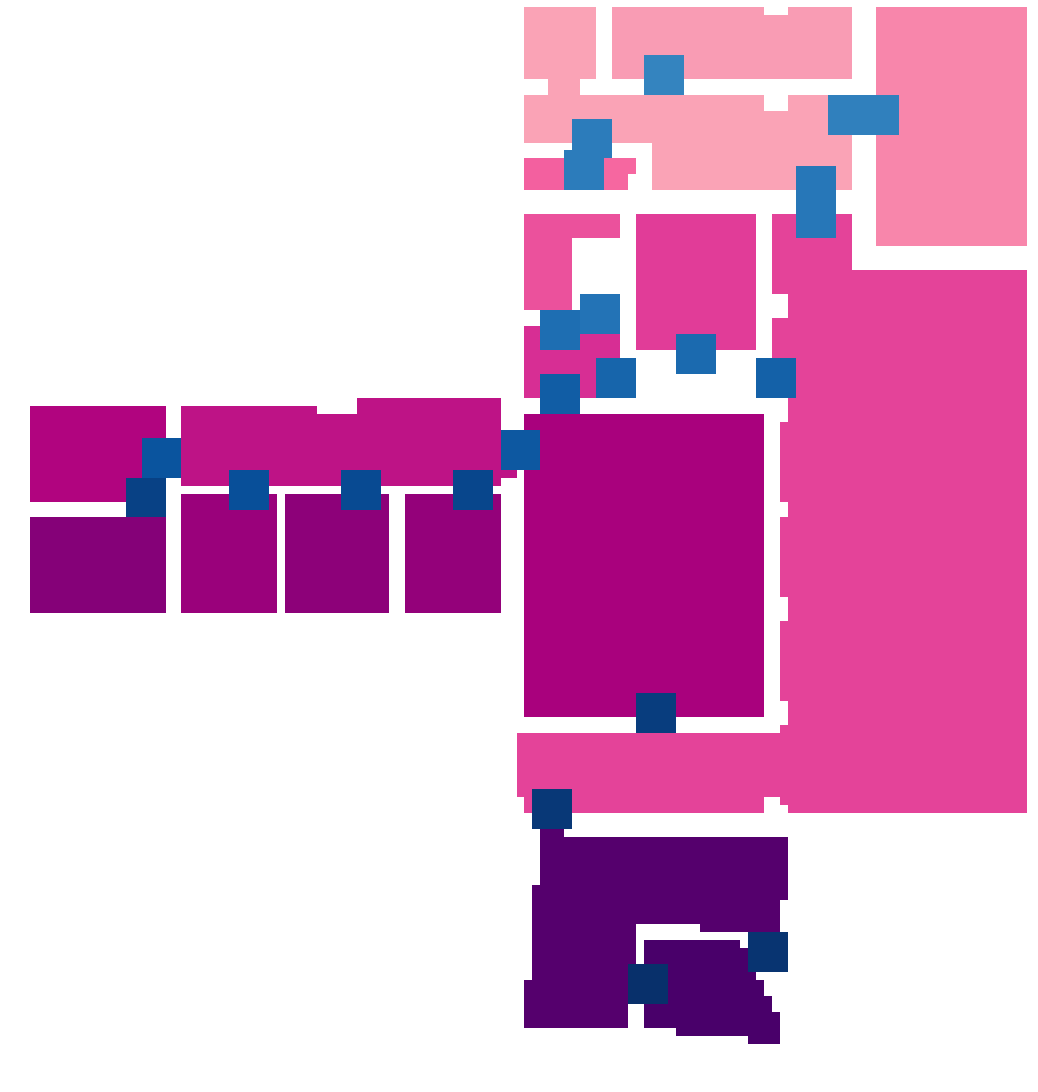}
		\label{fig:seg}
	}
	\caption{(a) Obtained poses of rooms and doors. The red point represent the poses of rooms. The blue points represent the poses of doors. (b) Created segmentation of the scene. The red and blue areas represent different rooms and doors respectively.}

	\vspace*{-5mm}
\end{figure}

Similarly with \cite{best2024multi}, the locations of rooms and doors that connect different rooms can be obtained by considering a 2D distance transform map of walls.
An illustration for this extraction can be found in Fig. \ref{fig:room_door}.
Then, we can obtain the segmentation of different rooms and doors using the flood fill algorithm.
The result can be found in Fig. \ref{fig:seg}.
Finally, the topology $\mathcal{G}$ between each room can be obtained.
In $\mathcal{G}$, the vertex set $\mathcal{V}$ consists of the central positions of each room, with an edge existing between two adjacent vertices. 
The weight of the edge is the Euclidean distance between the centers of the two rooms.

In the following sections, we will demonstrate how to utilize the predicted segmentation and topological connectivity to accelerate exploration and navigation tasks.

\subsubsection{Exploration Strategy}
\label{sec:exploration_strategy}
%Based on the method of predicting the floor plan from a noisy 2D grid map, we apply it to the exploration task in unknown environments. 

Firstly, at time step $t$, we assume that the robot is at $p(t)$ and the constructed room topology is $\mathcal{G}(t)$.
The state of each room can be classified based on whether there are frontiers within the room. 
We assume that the set representing rooms containing frontiers is denoted as $\mathcal{V}_{f,i}(t),i=1,\cdots,N_i$.
$N_i$ is the number of rooms that has frontiers.
In indoor environments, the robot needs to fully explore one room before moving to another \cite{kim2023multi} to accelerate exploration.
Therefore, optimizing the room visiting order will benefit this task.
Firstly, we obtain the visiting order of each room including the predicted ones in $\mathcal{V}_{f,i}(t)$ by minimizing the total path length.
It can be formulated as a linear programming problem
\begin{equation}
    \begin{split}
	\min & \sum_{i \in \mathcal{U}} \sum_{i\neq j, j \in \mathcal{U}}  c_{ij} z_{ij}\\
        s.t. \ &z_{ij}\in \{0,1\}, i,j \in \mathcal{U},i \neq j\\
        &\sum_{i\in \mathcal{U}, i\neq j} z_{ij} = 1,j\in \mathcal{U} \backslash q(t)\\
        &\sum_{j\in \mathcal{U} \backslash q(t)} z_{ij} \leq 1, i\in \mathcal{U}\\
        &z_{q(t)j} = 1\\
        &\sum_{i\in \mathcal{S}} \sum_{j\in \mathcal{S},i\neq j} z_{ij} \leq |S| - 1, \forall \mathcal{S} \subset \mathcal{U}, |\mathcal{S}|>1.
    \end{split}
\end{equation}
We define $\mathcal{U} = \{\mathcal{V}_{f,i}(t),i=1,\cdots,N_i\} \cup \{p(t)\}$ which adds the robot pose into the set of rooms with frontiers. $\mathcal{U}$ indicates all nodes involved in this problem.
$z_{ij}$ defines the existence of a certain path from $i$ to $j$, and $z_{ij}=1$, if it exists in the optimal solution.
The first equality indicates that all rooms should be visited one time.
The first inequality indicates the robot should leave each room at most one time.
The last inequality implies that there are no loops in the optimal path.
This problem can be solved using Lin-Kernighan heuristic algorithm\cite{helsgaun2000effective}.

Therefore, we can obtain the visiting order of each room denoted as $u_i,1\leq u_i \leq N_i$.
We use the visiting order of rooms, which includes the predicted information, to provide a robust guidance for exploration.
The robot tends to select its next target considering both $u_i$ which contains the predicted information and path length which contains the accurate information.
The cost function for the next target is
\begin{equation}
	C(f_i) = \exp(\lambda u_j)  C_N(f_i,p(t)),f_i\in \mathcal{V}_{f,j},
	\label{equ:explore}
\end{equation}
where $\lambda$ is a hyper-parameter that indicates the impact of the visiting order.
$\exp(\lambda u_j)$ represents the influence of the visit order on the selection of the next target.
A larger $\lambda$ means that the exploration strategy tends to rely more on the predicted room locations and topological connectivity as guidance. 
$C_N(f_i,p(t))$ is the navigation cost between $f_i$ and $p(t)$.
The robot will choose the frontier with minimal $C(f_i)$ as its next target.
In section \ref{sec:th}, we will discuss the influence of $\lambda$ on the exploration result.

Compared to strictly requiring the robot to explore according to the optimized room visiting order, this formula employs a soft constraint, thus achieving a balance between predicted information and current observations. 
This ultimately improves the robustness of the exploration performance with respect to the map predictor.

\section{Experiments}
We study the application of floor plan prediction in exploration. 
In simulations, a robot is equipped with a 12\,m range LiDAR. 
Besides, the localization system is assumed to be perfect, which is similar to \cite{zhou2023racer}.
Training of the NN and simulations are conducted using AMD Ryzen 3900X CPU, 64\,GB RAM, and NVIDIA RTX 3090 GPU.
In Eq. \ref{equ:loss}, the weight for occupied space is 0.6, and the weights for free and unknown spaces are 0.2.
In Eq. \ref{equ:explore}, $\lambda$ is set to 0.08.

\subsection{Comparison of Map Prediction Results}

\begin{figure*}[!t]
	\centering  
	\includegraphics[width=6.5in]{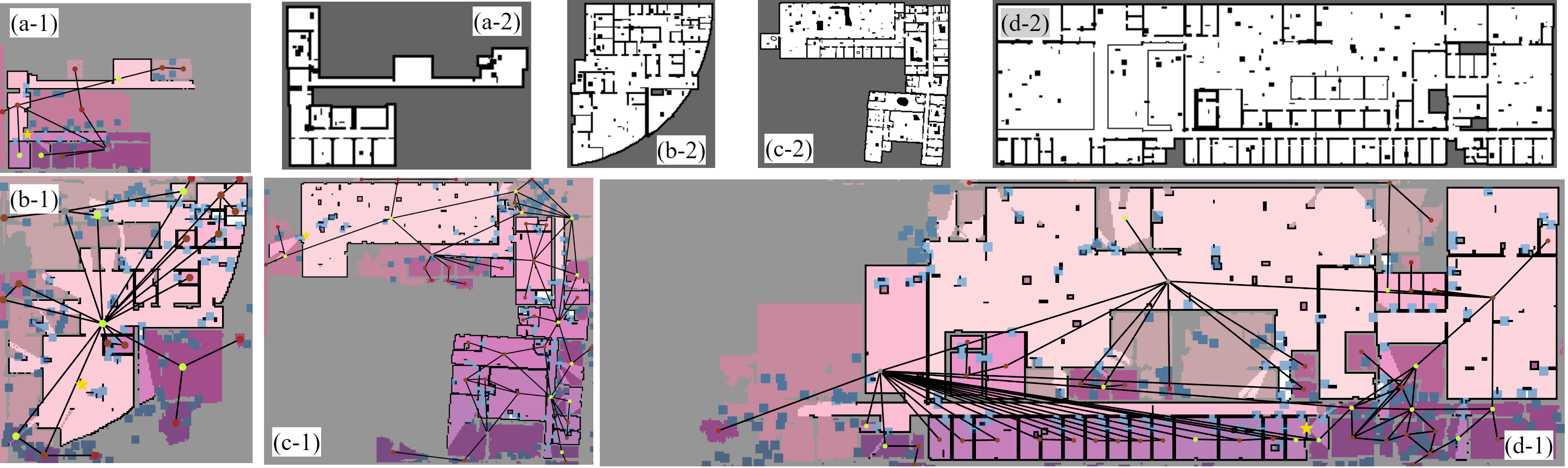}
	\caption{Illustration for map prediction in different scenes. 
		Grid map: observations; Masks: predicted maps.
		The first picture in each group is the predicted floor plan. 
		The second picture is the ground truth scene. 
		The scene is getting larger from (a) to (d).}
	\label{fig:case_study}
\end{figure*}

The qualitative results of the proposed method for prediction in different scenes can be found in Fig. \ref{fig:case_study}.
We choose four typical scenes in the test set for evaluation.
The areas of four scenes are 228.52, 604.84, 1126.44, and 2577.4\,m$^2$.
In (a), the scene is relatively small, and only a few rooms are in the scene.
Our methods can predict the existence of unseen rooms, while the detailed shapes are inaccurate.
For (b) and (c), there are multiple small rooms in the scene.
(d) has a complex structure.
Overall, our method can make reasonable predictions in various scenes, and the predicted topology can represent the structures of the scenes.

\begin{table}[t]
	\centering
	\fontsize{8}{11}\selectfont
	\caption{Comparison of Map Prediction Results}
	\vspace{-2mm}
	\label{table:pre_res}
	\begin{threeparttable}
		\begin{tabular}{@{}c@{\hspace{3pt}}c@{\hspace{3pt}}c@{\hspace{3pt}}c@{}c@{}c@{}}
			\toprule
                \toprule
			\multirow{2}{*}{\textbf{Methods}}& \multicolumn{1}{c}{$\rm{\textbf{Recall}}$} &\multicolumn{1}{c}{$\rm{\textbf{Precision}}$}&\multicolumn{1}{c}{$\rm{\textbf{F1\ Score}}$}&\multicolumn{1}{c}{${\rm{\textbf{Params}}}$}&\multicolumn{1}{c}{${\rm{\textbf{MACs}}}$}\\
                &(\%)&(\%)& &(M)&(G)\\
			\midrule
                VAE\cite{shrestha2019learned} &  41.5   & 60.7 & 0.493 & 27.7 & 6.01  \\
                ViT \cite{dosovitskiy2020image} & 63.4  & 52.2 & 0.576 & 80.7 & 3.51  \\
			UNet \cite{ronneberger2015u} & \textbf{80.4} & 70.1  & 0.749 & 7.76 & 3.43 \\
			UNet++\cite{zhou2018unet++} & 78.5 &  76.0 & 0.772 & 11.8 & 12.5 \\
			FPUNet &  76.1&  {\textbf {84.6 }}  & {\textbf{ 0.801}} & 20.7 & 14.6 \\
			\bottomrule
                \bottomrule
		\end{tabular}
	\end{threeparttable}
	\vspace{-2mm} 
\end{table}

To provide a quantitative comparison with other methods, we compare our FPUNet with other architectures of neural networks.
In \cite{shrestha2019learned}, Variational Autoencoders (VAE) with the ResNet \cite{he2016deep} architecture is used as the map predictor.
Vision Transformer (ViT)\cite{dosovitskiy2020image} uses attention to capture information in the images, which is the SOTA network for some vision tasks.
UNet \cite{ronneberger2015u} and UNet++ \cite{zhou2018unet++} are two popular frameworks for some tasks like segmentation.
The implementation of VAE including loss function, input, and output follows \cite{shrestha2019learned}.
The other NNs follow the same implementation of FPUNet.
Due to the position of walls being more important for prediction, we use three metrics of walls for evaluation.
The detailed results are presented in Table \ref{table:pre_res}.

In this task, recall and precision are two conflicting metrics. 
A higher recall means that the NN makes more predictions, which inevitably leads to more incorrect predictions, thereby lowering precision. 
Among all the models, UNet achieves the best recall at 80.4\%, but its precision was only 70.1\%, indicating a large number of incorrect predictions. 
FPUNet achieved the best precision, meaning its predictions are the most accurate. 
We use the F1 score to provide a comprehensive evaluation of the NN's performance. 
FPUNet achieves the best result in terms of the F1 score.
In addition, higher precision is also beneficial for subsequent exploration tasks. 
Since the NN provides more reliable prediction information, it can offer effective guidance for exploration.

\subsection{Exploration Results}
\label{sec:exp}
Out of all 140 scenes, 21 scenes are selected as the test set. 
We study exploration efficiency in small-sized (smaller than 200 m$^2$), medium-sized (200–600 m$^2$), and large-sized scenes (greater than 600 m$^2$) in the test set.
The quantities of these three types are 6, 7, and 8, respectively.

We compare \textit{P$^2$ Explore} with three different exploration strategies.
The first one is NBV \cite{umari2017autonomous}, where the next goal is selected as the frontier with maximum utility.
The second one is TSP\cite{zhou2023racer,hardouin2023multirobot}, which can be seen as the SOTA traditional method.
In TSP, the robot considers the distribution of all frontiers during each step and uses TSP algorithm to traverse all the frontiers.
These two methods are algorithms based on current observation information. 
We also compare \textit{P$^2$ Explore} with strategies based on predictions \cite{shrestha2019learned}. 
However, in Pre-based \cite{shrestha2019learned}, the VAE-based map predictor performed poorly in cluttered environments. 
Therefore, we use the FPUNet proposed in this paper as its map predictor. 
\cite{shrestha2019learned} acquires predicted information gain based on the predicted map, and selects the next exploration frontier point based on the predicted information gain and path cost.

\begin{table}[!t]
	\fontsize{8.5}{7}\selectfont
	\renewcommand{\arraystretch}{1.8} 
	\setlength\tabcolsep{2pt}   
	\caption{Comparison and ablation studies of exploration efficiency. 
		Evaluation metric: path length (m).}
	\label{table:compare_exp}
	\centering
	\begin{tabular}{C{2cm}C{2cm}C{2cm}C{2cm}}
		\toprule
		\toprule
		\multicolumn{1}{c}{\footnotesize \textbf{Method}} & \multicolumn{1}{c}{\makecell{\footnotesize Small\\ ($<$200\,m$^2$)}} &\multicolumn{1}{c}{\makecell{\footnotesize Middle\\ (200-600\,m$^2$)}}   &  \multicolumn{1}{c}{\makecell{\footnotesize Large\\ ($>$600\,m$^2$)}}  \\
		\cmidrule(lr){1-4}
		NBV \cite{umari2017autonomous} & 87.20 (\scriptsize 42.54) 	&379.55 (\scriptsize 110.12) &948.34 (\scriptsize 486.97) \\
		TSP \cite{zhou2023racer}&  71.77 (\scriptsize 38.85)	&261.65 (\scriptsize 94.18) &652.31 (\scriptsize 402.99)\\
		Pre-based\cite{shrestha2019learned}&  62.50 (\scriptsize 27.76)& 255.08 (\scriptsize 81.76) & 654.76 (\scriptsize 339.06)\\
		w/o FPUNet&  63.57 (\scriptsize 33.30)& 253.04 (\scriptsize 87.86) & 655.49 (\scriptsize 334.02)\\
		\textit{P$^2$ Explore}&   \textbf{58.71} (\scriptsize 29.16)	&\textbf{249.51} (\scriptsize 81.71) &\textbf{620.19} (\scriptsize 287.25) \\
		\bottomrule
		\bottomrule
	\end{tabular}
\end{table}

The detailed results can be found in Table \ref{table:compare_exp}.
In all three different types of scenarios, TSP shows better exploration efficiency than NBV. 
Especially in large scenes, TSP achieves a 31.21\% reduction in path length due to it optimizes the visiting order of current observations.
Compared with TSP, \textit{P$^2$ Explore} achieve improvements of 18.20\%, 4.63\%, and 4.92\% in three different sizes of scenes, respectively.
\cite{shrestha2019learned} introduces predicted information for the NBV exploration strategy. 
It can be observed that by utilizing the predicted information, the exploration path is reduced by 24.7\,m, 124.47\,m, and 293.58\,m on average, respectively.
However, even though we equip \cite{shrestha2019learned} with the same map predictor which is better than its original one, \textit{P$^2$ Explore} demonstrates some advantage. 
The length of our exploration path is reduced by 3.79\,m, 5.57\,m, and 34.57\,m in the three scenarios, respectively.
Especially in larger scenarios, our method achieved a 5.28\% improvement. 
This improvement comes from our method's incorporation of the advantages of the TSP, using the room visiting order as guidance. 
This allows the predicted information to be applied more robustly and also enables the planning of a globally optimal trajectory.

To demonstrate that our improvement comes from the introduction of predictive information, we conduct an ablation study. 
In \textit{w/o FPUNet}, we remove the predictive information while keeping all other components unchanged. 
The results show that \textit{P$^2$ Explore} enhances exploration efficiency by incorporating predictive information. 
Especially in larger environments, the use of predictive information leads to even greater improvements.

\subsection{Threshold Discussion}
\label{sec:th}
\subsubsection{Selection of weight for loss function}
\begin{figure}[!t]
	\centering  
	\subfloat[]{ 
		\centering    
		\includegraphics[width=1.6in]{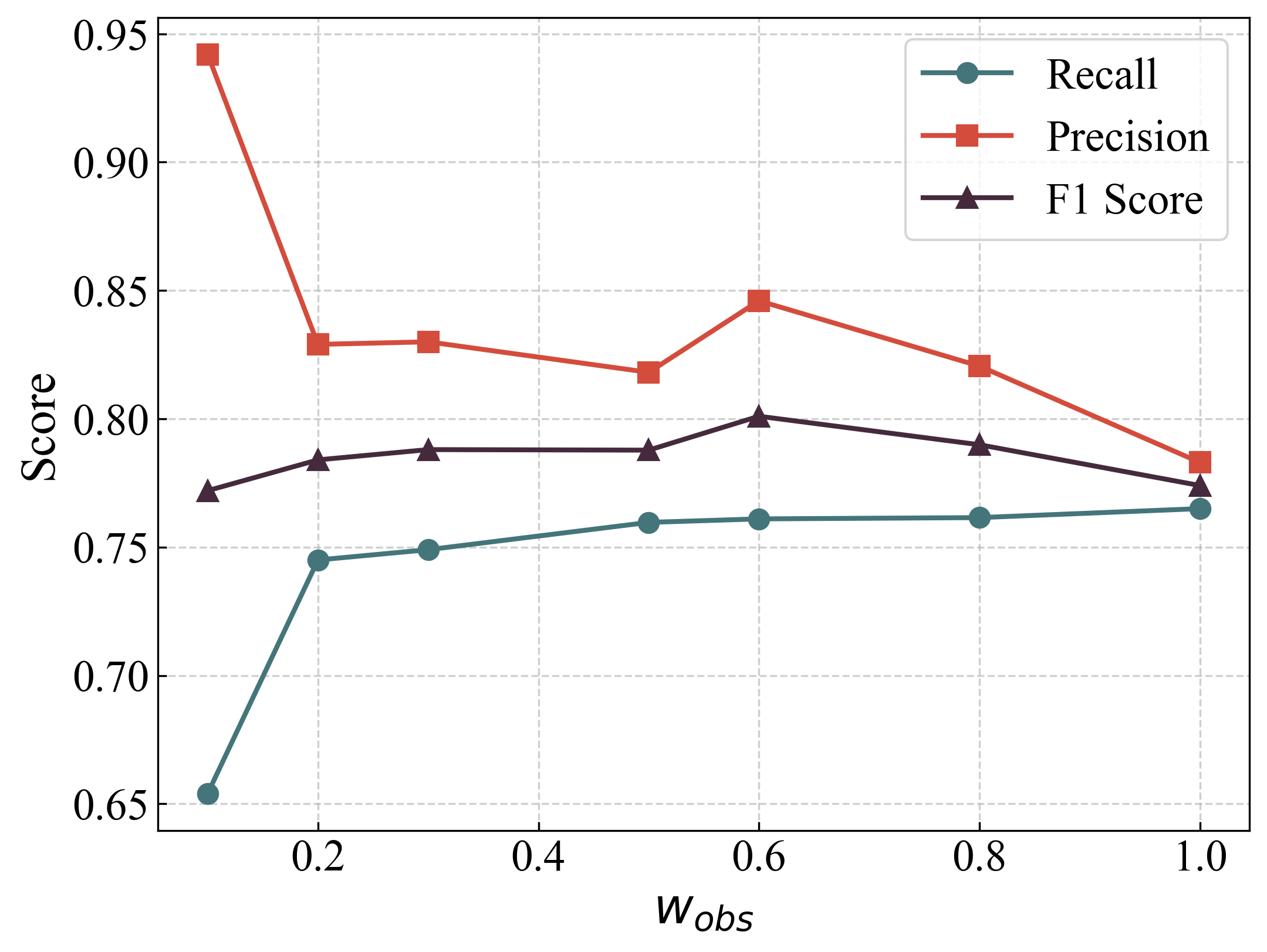}
		\label{fig:w_dis}
	}
	\subfloat[]{   
		\centering    
		\includegraphics[width=1.6in]{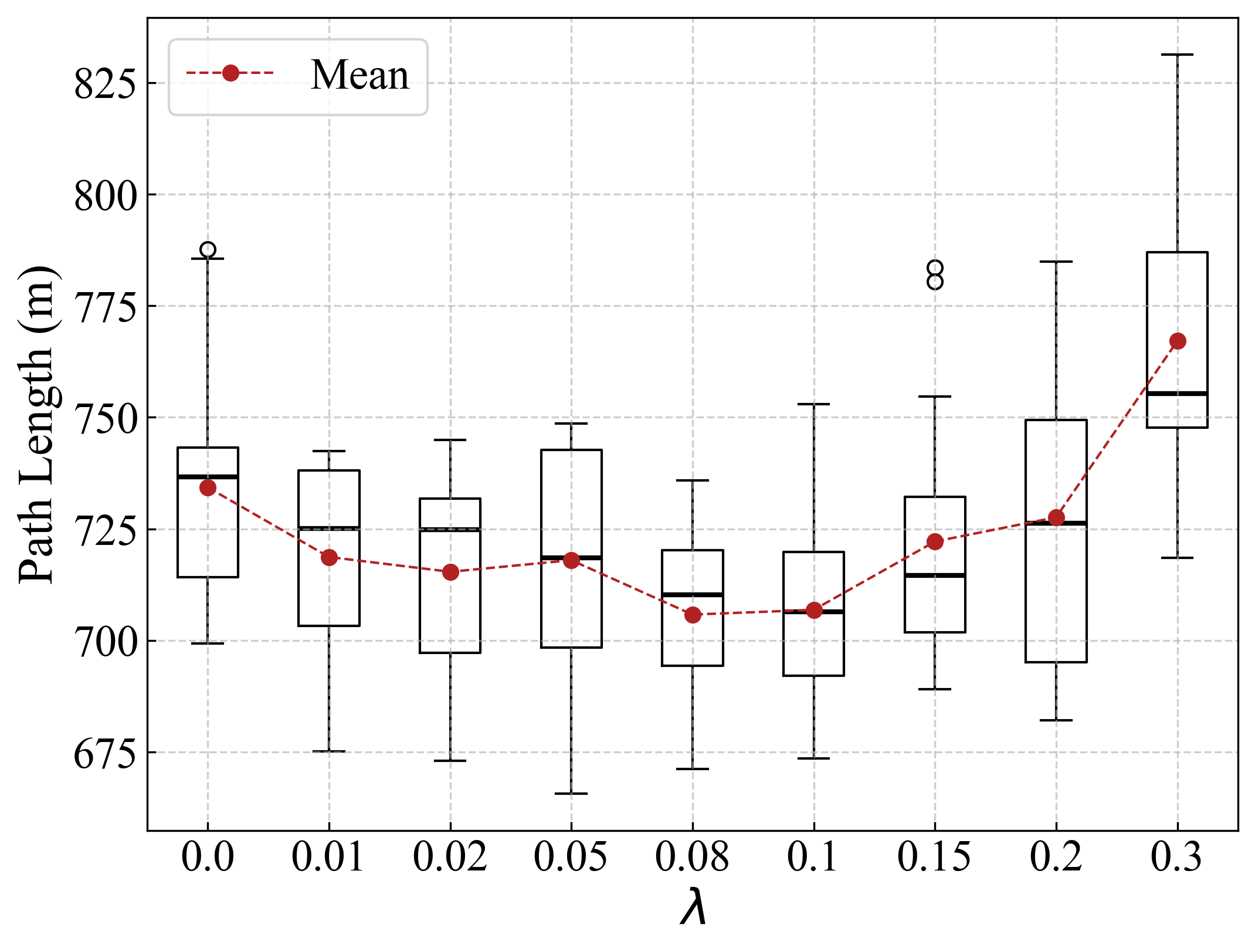}
		\label{fig:lamda_dis}
	}
	\caption{(a) The influence of obstacle weight for map prediction. (b) The influence $\lambda$ for exploration path length.}
	\label{fig:Compare_exp}
	\vspace*{-5mm}
\end{figure}

We investigate the impact of obstacle weight $w_{obs}$ in the loss function on the final prediction results. 
The weights of other two categories are set to a fixed value of 0.2. 
The results are shown in Fig. \ref{fig:w_dis}. 
It can be observed that as $w_{obs}$ increases, the recall gradually increases. 
This is because a larger weight makes the NN more likely to predict unknown obstacle structures. 
Precision, however, gradually decreases as the weight increases. 
This is mainly due to the model making more predictions, which leads to a higher number of incorrect predictions.
The F1 score is the average of recall and precision, and it yields better results when an appropriate weight is set. 
It can be observed that the optimal value is achieved around 0.6.

In addition, we also observed that even without selecting the optimal parameters, FPUNet can still achieve SOTA performance in terms of the F1 score. 
When $w_{obs}$ is set to 0.5 and 0.8, the F1 scores are 0.788 and 0.790, respectively, which outperforms UNet++'s score of 0.772.

\subsubsection{Selection of parameter for exploration}
In section \ref{sec:exploration_strategy}, we use a hyper-parameter $\lambda$ to control the influence of predicted information in exploration.
We study the impact of $\lambda$ on the path length and select a representative scene from the test set, with an area of 1121.92 m$^2$. 
In this scene, we choose 15 different starting points and test each parameter at these 15 points. 
The results are shown in the Fig. \ref{fig:lamda_dis}.

When $\lambda$ is set to a reasonable value, such as around 0.08, our method achieves the best performance. 
If a very small value, such as 0, is chosen, the exploration strategy tends to select the nearest frontier for exploration. 
This leads to an insufficient usage of predicted information and room visiting order, thus reducing the exploration efficiency. 
On the other hand, if a value that is too large is chosen, the robot becomes overly sensitive to incorrect predictions, like the incorrect existence of predicted rooms and their topology.
Therefore, the ignorance of the right information from current observations, like the path lengths to frontiers, leads to the reduction of efficiency.

\subsection{Real-world Experiments}
Since the training set is based on the KTH floor plan dataset, the effectiveness and generalization performance remain in doubt in other scenes. 
Therefore, we validate our proposed method in a real-world laboratory scene. 
We first use a Turtlebot equipped with an RPLIDAR A2 LIDAR (range 12\,m) to reconstruct the environment into a 2D grid map using Cartographer.
Note that we do not deploy the entire \textit{P$^2$ Explore} on the real-world robot, primarily due to insufficient hardware computing power, which prevents us from running the NN inference.
Therefore, we perform simulated exploration in the obtained 2D map.
\begin{figure}[!t]
	\centering    
	\includegraphics[width=3.4in]{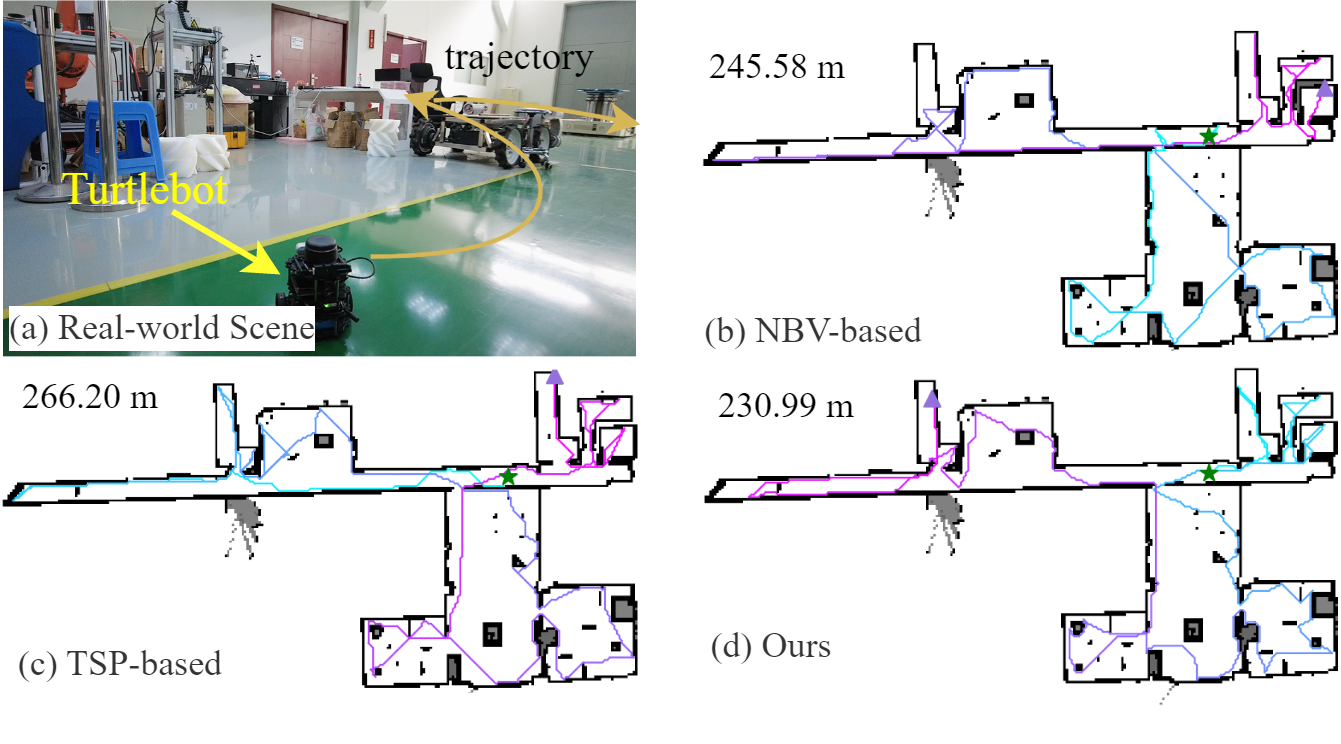}
	\caption{Illustration of the real-world experiment.
		(a) A screenshot captured of the robot in the real-world scene.
		(b) (c) (d) Results of different methods in the scene.
	}
	\label{fig:realworld}
	\vspace*{-5mm}
\end{figure}

The result can be found in Fig. \ref{fig:realworld}.
The effectiveness of the proposed method is compared with \textbf{NBV} and \textbf{TSP}.
The total path lengths of \textbf{NBV} and \textbf{TSP} are 245.58\,m, and 266.20\,m, respectively.
By implementing map prediction in the process, the robot uses 230.99\,m to finish the exploration.
This result indicates that map prediction can exhibit the power of generalization and accelerate exploration, even in previously unseen scenarios.

\section{Conclusion \& Future Work}
In this work, we focus on facilitating the efficiency of exploration using map prediction in cluttered environments.
Firstly, we perform floor plan prediction using FPUNet which is the SOTA method for this task.
Additionally, we also extract the positions of rooms and their connectivity based on the predicted results. 
The optimized visiting order of rooms is used to provide a high-level guidance for exploration.
The effectiveness of the proposed method is demonstrated by exploration in unknown environments.
Our method can shorten the path length compared with the various baselines.
Currently, the algorithm has not been deployed in a real robot. 
In future work, more comprehensive experimental validation can be conducted to verify its feasibility in a real-world setting.
Besides, lightweight network architectures can be considered to enable map prediction.

\bibliographystyle{ieeetr} 
\bibliography{ref}

\begin{thebibliography}{10}

\bibitem{hoeller2024anymal}
D.~Hoeller, N.~Rudin, D.~Sako, and M.~Hutter, ``Anymal parkour: Learning agile
  navigation for quadrupedal robots,'' {\em Sci. Robot.}, vol.~9, no.~88,
  p.~eadi7566, 2024.

\bibitem{yokoyama2024vlfm}
N.~Yokoyama, S.~Ha, D.~Batra, J.~Wang, and B.~Bucher, ``Vlfm: Vision-language
  frontier maps for zero-shot semantic navigation,'' in {\em Proc. IEEE Int.
  Conf. Robot. Autom.}, pp.~42--48, IEEE, 2024.

\bibitem{padalkar2023open}
A.~Padalkar, A.~Pooley, A.~Jain, A.~Bewley, A.~Herzog, A.~Irpan, A.~Khazatsky,
  A.~Rai, A.~Singh, A.~Brohan, {\em et~al.}, ``Open x-embodiment: Robotic
  learning datasets and rt-x models,'' {\em arXiv preprint arXiv:2310.08864},
  2023.

\bibitem{cao2023representation}
C.~Cao, H.~Zhu, Z.~Ren, H.~Choset, and J.~Zhang, ``Representation granularity
  enables time-efficient autonomous exploration in large, complex worlds,''
  {\em Sci. Robot.}, vol.~8, no.~80, 2023.

\bibitem{zhou2023racer}
B.~Zhou, H.~Xu, and S.~Shen, ``Racer: Rapid collaborative exploration with a
  decentralized multi-uav system,'' {\em {IEEE} Trans. Robotics}, 2023.

\bibitem{best2024multi}
G.~Best, R.~Garg, J.~Keller, G.~A. Hollinger, and S.~Scherer, ``Multi-robot,
  multi-sensor exploration of multifarious environments with full mission
  aerial autonomy,'' {\em Int. J. Robot. Res.}, vol.~43, no.~4, pp.~485--512,
  2024.

\bibitem{tao2023seer}
Y.~Tao, Y.~Wu, B.~Li, F.~Cladera, A.~Zhou, D.~Thakur, and V.~Kumar, ``Seer:
  Safe efficient exploration for aerial robots using learning to predict
  information gain,'' in {\em Proc. IEEE Int. Conf. Robot. Autom.},
  pp.~1235--1241, IEEE, 2023.

\bibitem{tao2024learning}
Y.~Tao, E.~Iceland, B.~Li, E.~Zwecher, U.~Heinemann, A.~Cohen, A.~Avni, O.~Gal,
  A.~Barel, and V.~Kumar, ``Learning to explore indoor environments using
  autonomous micro aerial vehicles,'' in {\em Proc. IEEE Int. Conf. Robot.
  Autom.}, pp.~15758--15764, IEEE, 2024.

\bibitem{katyal2019uncertainty}
K.~Katyal, K.~Popek, C.~Paxton, P.~Burlina, and G.~D. Hager,
  ``Uncertainty-aware occupancy map prediction using generative networks for
  robot navigation,'' in {\em Proc. IEEE Int. Conf. Robot. Autom.},
  pp.~5453--5459, IEEE, 2019.

\bibitem{elhafsi2020map}
A.~Elhafsi, B.~Ivanovic, L.~Janson, and M.~Pavone, ``Map-predictive motion
  planning in unknown environments,'' in {\em Proc. IEEE Int. Conf. Robot.
  Autom.}, pp.~8552--8558, IEEE, 2020.

\bibitem{wei2021occupancy}
M.~Wei, D.~Lee, V.~Isler, and D.~Lee, ``Occupancy map inpainting for online
  robot navigation,'' in {\em Proc. IEEE Int. Conf. Robot. Autom.},
  pp.~8551--8557, IEEE, 2021.

\bibitem{zheng2025aage}
L.~Zheng, M.~Wei, R.~Mei, K.~Xu, J.~Huang, and H.~Cheng, ``{AAGE}: Air-assisted
  ground robotic autonomous exploration in large-scale unknown environments,''
  {\em {IEEE} Trans. Robotics}, 2025.

\bibitem{hardouin2023multirobot}
G.~Hardouin, J.~Moras, F.~Morbidi, J.~Marzat, and E.~M. Mouaddib, ``A
  multirobot system for 3-d surface reconstruction with centralized and
  distributed architectures,'' {\em {IEEE} Trans. Robotics}, 2023.

\bibitem{umari2017autonomous}
H.~Umari and S.~Mukhopadhyay, ``Autonomous robotic exploration based on
  multiple rapidly-exploring randomized trees,'' in {\em Proc. IEEE/RSJ Int.
  Conf. Intell. Robot. Syst.}, pp.~1396--1402, 2017.

\bibitem{patwardhan2024distributed}
A.~Patwardhan and A.~J. Davison, ``A distributed multi-robot framework for
  exploration, information acquisition and consensus,'' in {\em Proc. IEEE Int.
  Conf. Robot. Autom.}, pp.~12062--12068, IEEE, 2024.

\bibitem{asgharivaskasi2024riemannian}
A.~Asgharivaskasi, F.~Girke, and N.~Atanasov, ``Riemannian optimization for
  active mapping with robot teams,'' {\em arXiv preprint arXiv:2404.18321},
  2024.

\bibitem{choset2000coverage}
H.~Choset, ``Coverage of known spaces: The boustrophedon cellular
  decomposition,'' {\em Auton. Robot.}, vol.~9, pp.~247--253, 2000.

\bibitem{shrestha2019learned}
R.~Shrestha, F.-P. Tian, W.~Feng, P.~Tan, and R.~Vaughan, ``Learned map
  prediction for enhanced mobile robot exploration,'' in {\em Proc. IEEE Int.
  Conf. Robot. Autom.}, pp.~1197--1204, IEEE, 2019.

\bibitem{ericson2021understanding}
L.~Ericson, D.~Duberg, and P.~Jensfelt, ``Understanding greediness in
  map-predictive exploration planning,'' in {\em Eur. Conf. Mob. Robots},
  pp.~1--7, IEEE, 2021.

\bibitem{katyal2021high}
K.~D. Katyal, A.~Polevoy, J.~Moore, C.~Knuth, and K.~M. Popek, ``High-speed
  robot navigation using predicted occupancy maps,'' in {\em Proc. IEEE Int.
  Conf. Robot. Autom.}, pp.~5476--5482, IEEE, 2021.

\bibitem{ramakrishnan2020occupancy}
S.~K. Ramakrishnan, Z.~Al-Halah, and K.~Grauman, ``Occupancy anticipation for
  efficient exploration and navigation,'' in {\em Proc. Eur. Conf. Comput.
  Vis.}, pp.~400--418, Springer, 2020.

\bibitem{wang2021learning}
L.~Wang, H.~Ye, Q.~Wang, Y.~Gao, C.~Xu, and F.~Gao, ``Learning-based 3d
  occupancy prediction for autonomous navigation in occluded environments,'' in
  {\em Proc. IEEE/RSJ Int. Conf. Intell. Robot. Syst.}, pp.~4509--4516, IEEE,
  2021.

\bibitem{sharma2023proxmap}
V.~D. Sharma, J.~Chen, and P.~Tokekar, ``Proxmap: Proximal occupancy map
  prediction for efficient indoor robot navigation,'' in {\em Proc. IEEE/RSJ
  Int. Conf. Intell. Robot. Syst.}, pp.~7135--7140, IEEE, 2023.

\bibitem{ji2023ddp}
Y.~Ji, Z.~Chen, E.~Xie, L.~Hong, X.~Liu, Z.~Liu, T.~Lu, Z.~Li, and P.~Luo,
  ``Ddp: Diffusion model for dense visual prediction,'' in {\em Proc. IEEE Int.
  Conf. Comp. Vis.}, pp.~21741--21752, 2023.

\bibitem{chen2023stexplorer}
B.~Chen, Y.~Cui, P.~Zhong, W.~Yang, Y.~Liang, and J.~Wang, ``Stexplorer: A
  hierarchical autonomous exploration strategy with spatio-temporal awareness
  for aerial robots,'' {\em ACM Trans. Intell. Syst. Technol.}, vol.~14, no.~6,
  pp.~1--24, 2023.

\bibitem{chang2007p}
H.~J. Chang, C.~G. Lee, Y.-H. Lu, and Y.~C. Hu, ``P-slam: Simultaneous
  localization and mapping with environmental-structure prediction,'' {\em
  {IEEE} Trans. Robotics}, vol.~23, no.~2, pp.~281--293, 2007.

\bibitem{luperto2019predicting}
M.~Luperto, V.~Arcerito, and F.~Amigoni, ``Predicting the layout of partially
  observed rooms from grid maps,'' in {\em Proc. IEEE Int. Conf. Robot.
  Autom.}, pp.~6898--6904, IEEE, 2019.

\bibitem{ericson2022floorgent}
L.~Ericson and P.~Jensfelt, ``Floorgent: Generative vector graphic model of
  floor plans for robotics,'' in {\em Proc. IEEE/RSJ Int. Conf. Intell. Robot.
  Syst.}, pp.~12485--12491, IEEE, 2022.

\bibitem{ericson2024beyond}
L.~Ericson and P.~Jensfelt, ``Beyond the frontier: Predicting unseen walls from
  occupancy grids by learning from floor plans,'' {\em {IEEE} Robot. Automat.
  Lett.}, 2024.

\bibitem{aydemir2012can}
A.~Aydemir, P.~Jensfelt, and J.~Folkesson, ``What can we learn from 38,000
  rooms? reasoning about unexplored space in indoor environments,'' in {\em
  Proc. IEEE/RSJ Int. Conf. Intell. Robot. Syst.}, pp.~4675--4682, IEEE, 2012.

\bibitem{zhou2018unet++}
Z.~Zhou, M.~M. Rahman~Siddiquee, N.~Tajbakhsh, and J.~Liang, ``Unet++: A nested
  u-net architecture for medical image segmentation,'' in {\em Proc. Int. Conf.
  Med. Image Comput. Comput.-Assist. Interv.}, pp.~3--11, Springer, 2018.

\bibitem{Li2022anisConvo}
J.~Li, P.~Wang, K.~Han, and Y.~Liu, ``Anisotropic convolutional neural networks
  for rgb-d based semantic scene completion,'' {\em {IEEE} Trans. Pattern Anal.
  Machine Intell.}, vol.~44, no.~11, pp.~8125--8138, 2022.

\bibitem{kim2023multi}
S.~Kim, M.~Corah, J.~Keller, G.~Best, and S.~Scherer, ``Multi-robot multi-room
  exploration with geometric cue extraction and circular decomposition,'' {\em
  {IEEE} Robot. Automat. Lett.}, 2023.

\bibitem{helsgaun2000effective}
K.~Helsgaun, ``An effective implementation of the lin--kernighan traveling
  salesman heuristic,'' {\em Eur. J. Oper. Res.}, vol.~126, no.~1,
  pp.~106--130, 2000.

\bibitem{dosovitskiy2020image}
A.~Dosovitskiy, ``An image is worth 16x16 words: Transformers for image
  recognition at scale,'' {\em arXiv preprint arXiv:2010.11929}, 2020.

\bibitem{ronneberger2015u}
O.~Ronneberger, P.~Fischer, and T.~Brox, ``U-net: Convolutional networks for
  biomedical image segmentation,'' in {\em Proc. 18th Int. Conf. Med. Image
  Comput. Comput.-Assist. Interv.}, pp.~234--241, Springer, 2015.

\bibitem{he2016deep}
K.~He, X.~Zhang, S.~Ren, and J.~Sun, ``Deep residual learning for image
  recognition,'' in {\em Proc. IEEE Conf. Comp. Vis. Pattern Recog.},
  pp.~770--778, 2016.

\end{thebibliography}
\end{document}